# Explainable AI in diagnosing and anticipating Leukemia using Transfer Learning method.

Wahidul Hasan Abir[1], Md. Fahim Uddin[1], Faria Rahman Khanam[1] and Mohammad Monirujjaman Khan[1,*]

[1]Department of Electrical and Computer Engineering,
North South University, Bashundhara, Dhaka-1229, Bangladesh
Mohammad Monirujjaman Khan. Email: monirujjaman.khan@northsouth.edu

**Abstract:** White blood cells (WBCs) are blood cells that fight infections and diseases as a part of the immune system. Hence, they are also known as "defender cells." But the imbalance in the number of WBCs in the blood can be hazardous. Leukemia is the most common blood cancer caused by an overabundance of white blood cells (WBCs) in the immune system. Acute lymphocytic leukemia (ALL) usually occurs when the bone marrow creates many immature WBCs that destroy healthy cells. Both children and teenagers are affected by acute lymphocytic leukemia (ALL). The rapid proliferation of atypical lymphocyte cells can cause a reduction in new blood cells and increase the chances of death in patients. Therefore, early and precise cancer detection can help with better therapy and a higher survival probability in the case of leukemia. However, diagnosing acute lymphocytic leukemia (ALL) is time-consuming and complicated, and manual analysis is expensive, with subjective and error-prone outcomes. Thus, detecting normal and malignant cells reliably and accurately is crucial. For this reason, automatic detection using computer-aided diagnostic (CAD) models can help doctors effectively detect early leukemia. The entire approach may be automated using image processing techniques, reducing physicians' workload and increasing diagnosis accuracy. Deep learning's impact on medical research has recently proven quite beneficial, offering new avenues and possibilities in the healthcare domain for diagnostic techniques. However, to make that happen soon in deep learning, the entire community must overcome the explainability limit. Because of the black box operation's shortcomings in AI models' decisions, there is a lack of liability and trust in the outcomes. But explainable artificial intelligence (XAI) can solve this problem by interpreting the predictions of artificial intelligence systems This paper focuses on leukemia, specifically Acute Lymphoblastic Leukemia (ALL). The proposed strategy recognizes acute lymphoblastic leukemia as an automated procedure that applies different transfer learning models to classify Acute Lymphoblastic Leukemia (ALL). Hence, using Local Interpretable Model-Agnostic Explanations (LIME) to assure validity and reliability, this method also explains the cause of a specific classification. The proposed method achieved 98.38% accuracy with the InceptionV3 model. Experimental results were found between different transfer learning methods, including ResNet101V2, VGG19, and InceptionResNetV2, later verified with the LIME algorithm for explainable AI, where the proposed method performed the best. The obtained results and its reliability demonstrate that it can be preferred in identifying Acute Lymphocytic Leukemia, which will assist medical examiners.

# 1 Introduction

Blood supplies essential substances to the entire human body. Erythrocytes (Red Blood Cells), leukocytes (White Blood Cells), and thrombocytes (Platelets) are the three main components of blood cells in the human body. Red Blood Cells (RBCs) ensure oxygen transportation to different parts of the human body, and in the case of injury, platelets help with blood clotting. White Blood Cells (WBCs) fight germs and prevent human infections. WBCs make up only 1% of blood volume, but slight changes are significant because the human immune system depends on WBCs. Any fluctuation in the number of leukocytes (WBCs) in the blood indicates a problem. Having an abnormally high number of WBC in our bodies can be detrimental and contribute to disease. Among them, leukemia is one of the most common diseases related to WBC count. [1] Leukemia is a prevalent and deadly disease. Leukemia is a cancer of leukocytes (WBCs) that affects the blood-forming cells. Many teenagers and children are at risk of developing leukemia. According to a 2012 study, around 352,000 people and children worldwide get leukemia, which begins inside the bone marrow and is identified by an unexpected growth in the number of white blood cells. [2] These defective blood cells put the immune system at risk, which affects the blood and bone marrow. Furthermore, these malignant WBCs can enter the bloodstream, and these cancerous cells can spread to multiple organs and harm the entire body via infected blood cells. which can be threatening if not diagnosed early or if therapy is delayed. [3]

Leukemia is classified primarily based on whether it is growing rapidly (acute) or slowly (chronic). Each of these types can be fatal if not detected or if therapy is delayed. Chronic leukemia usually takes a long time to develop. In contrast, the average survival time for acute leukemia patients without specific treatment is only three months. Acute lymphocytic leukemia is the most common among children, accounting for 25% of all childhood cancers. [4] Acute Lymphoblastic Leukemia (ALL) can lead a patient to death. If it is detected early on, it is generally treatable, and the patient's chances of survival increase. That is why early detection of immature cell formations is necessary to increase the patients' survival rate. Early and accurate diagnosis could help patients save money on therapy and increase their chances of remission. The limitations of diagnosing leukemia patients by humans are time-consuming and can become error-prone. An inaccurate diagnosis can threaten a patient's health. And in addition to making treatment more difficult, it raises treatment expenses. Hence, developing automated, low-cost systems that can accurately identify healthy and abnormal blood smear images is crucial. Many assist systems have been proposed to assist physicians in achieving high diagnosis accuracy. Physicians can diagnose a disease based on a specific dataset, including signs, symptoms, medical images, and exams.

Many researchers have proposed many strategies and algorithms for recognizing, segmenting, and classifying Acute Lymphoblastic Leukemia (ALL). The success of classification is dependent on the success of feature extraction, which is dependent on the success of segmentation. Hence, high classification accuracy requires the execution of all procedures. Deep learning has recently achieved remarkable progress in computer vision, image processing, and recognition. It has become a promising choice for medical image analysis. Among them, a considerable amount of work has been focused on leukemia diagnosis. [5] For example, Nayaki et al. employed a deep learning system based on image processing methods and a CNN to detect defective blood cells in microscopic blood images and achieved an accuracy rate of 80.4%. [6] Kasani et al. presented a study to classify leukemic B-lymphoblast for the development of an aggregated deep learning model. A trustworthy and accurate deep learner was created that can correctly diagnose acute lymphoblastic leukemia with a 96.58% classification accuracy using a small dataset size. [7] Some researchers use the Convolutional Neural Network (CNN) method to diagnose leukemia. CNN is the most extensively used method for image recognition. It has high self-learning, adaptability, and generalization abilities. Traditional image recognition methods need feature extraction and classification, whereas CNN requires only the image data as an input to complete the image classification with the network's self-learning ability. [8] Genovese et al. have introduced the first method for ALL detection based on histopathological transfer learning. On a histopathology database, CNN is trained before being fine-tuned on the ALL database to recognize lymphoblast tissue types with an accuracy rate of 88.69%. [9] Similarly, Safuan et al. applied classified the WBC types to identify

Acute Lymphoblastic Leukemia (ALL) with CNN, where pre-trained models of deep learning like AlexNet, GoogleNet, and VGG-16 are differentiated from each other to find the model that can classify better with a classification accuracy rate of 96.15%. [10] Jian et al. have proposed the ViT-CNN ensemble model to help diagnose acute lymphoblastic leukemia by classifying cancerous and normal cells. The ViT-CNN ensemble model extracts features from cell pictures in two alternative ways to improve classification results in a very accurate detection method that reaches 99.03% accuracy. [11] Aftab et al. have proposed a methodology for detecting leukemia using the Apache Spark BigDL library and Convolutional Neural Network (CNN) architecture GoogleNet deep transfer learning using microscopic images of human blood cells and reached a 96.42% accuracy rate. [12] However, the lack of explainability of neural networks limits the wide-scale adoption of deep learning in healthcare applications where explaining the fundamental logic is vital for decision-makers. Machine learning models are often contemplated to be "black boxes" that are tough to decipher. Among them, neural networks used in deep learning are the most difficult to comprehend. [13] If AI can't explain itself in the healthcare domain, the risk of making a wrong decision may outweigh the benefits of precision, speed, and decision-making efficacy. As a result, its scope and utility would be severely limited. That is why Explainable AI (XAI) can better understand and explain deep learning and neural networks. The adoption of Explainable AI (XAI) techniques is justified by the desire to promote transparency, result tracking, and model improvement. For example, Pawar et al. discussed XAI as a technique for AI-based systems to analyze and diagnose health data to provide accountability, transparency, result tracking, and model improvement in the healthcare domain. [14] Besides, Arrieta et al. proposed an analysis of recent contributions towards the explainability of different machine learning models focused on explaining various deep learning methodologies. [15] These intelligent healthcare systems can then be utilized to diagnose leukemia and choose the best treatment option. Explainable AI (XAI) can explain both the diagnosis result and the radix of the prediction. Deep neural network construction and training are time-consuming and very complex processes. So, instead of creating a deep neural network from scratch, the concept of transfer learning can be applied, where a deep network that has successfully solved one problem is customized to solve another.

The proposed approach in this paper aims to implement and compare different transfer learning models of TensorFlow in classifying Acute Lymphocytic Leukemia (ALL) which will help doctors to detect Acute Lymphocytic Leukemia (ALL) cells in patients to save human lives. By comparing different transfer learning models, future research on Acute Lymphocytic Leukemia (ALL) classification will get a headstart in choosing transfer learning models. Also, this method manages to explain which part of the image from the dataset caused the model to make a specific classification using Local Interpretable Model-Agnostic Explanations (LIME) to assure the model's validity and reliability. As deep learning in terms of medical classification is getting more popular, it is very important to know the cause of a prediction so that the doctor can easily verify the result. This procedure will make Acute Lymphocytic Leukemia (ALL) classification easier, more accurate, and more reliable.

The proposed method compares different transfer learning methods with high accuracy and an f1 score, which can identify Acute Lymphocytic Leukemia (ALL). The proposed method uses stratified KFold and uses explainable AI to showcase the image concentration of sample cells, which is novel in classifying leukemia with high precision. Usage of explainable AI makes the proposed method very reliable for medical examiners. It reduces costs and saves time by automating the process. Also, comparison between different transfer learning methods helps future researchers to choose suitable methods for leukemia research, which paves the way for further improvement.

The remaining contents of the paper are arranged as follows: Section 2 proposes the method and materials. This section gives a gist of the system model and the whole system. Then, results and analysis are described in Section 3. Lastly, in Section 4, the paper ends with the conclusion.

## 2 Materials and Methodology

### 2.1 Dataset Description

The dataset used for leukemia cancer identification is obtained from the CNMC website, which was used for the Acute Lymphocytic Leukemia (ALL) challenge of ISBI 2019. [16]

These cells were segmented from microscopic images and are indicative of real-world photos since they contain staining noise and illumination flaws, but these faults were mostly addressed after acquisition. This is because morphological similarities make it difficult to distinguish young leukemic eruptions from normal cells under a microscope. An expert oncologist annotated the ground truth labels. There are 15,114 photos in all from 118 patients, divided into two classes: normal cell and leukemia blast cell. The photos in the dataset are 450px*450px and in bmp format. In the following **Fig. 1**, there are sample images from two classes.

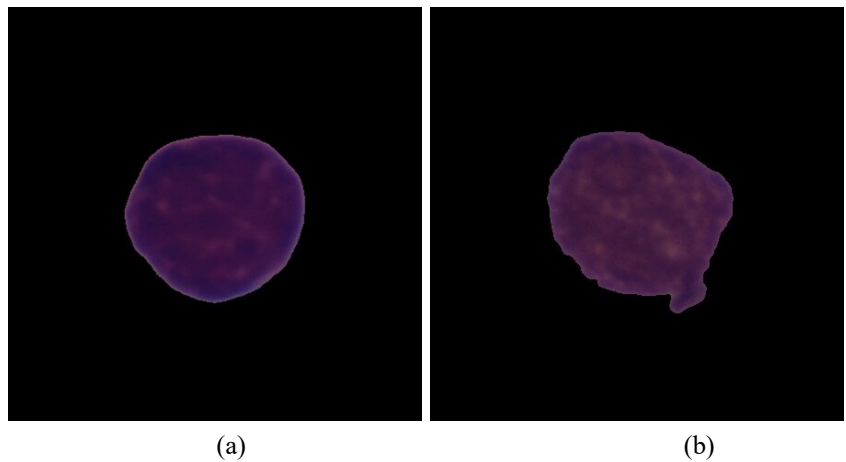

(a)  (b)

**Figure 1:** (a) Normal cell (b) Leukemia blast cell

### 2.2 Proposed Framework

The focus of this method is to detect acute lymphocytic leukemia by using transfer learning and later validate the model using explainable AI. Firstly, the leukemia images were collected from the CNMC website. As the dataset was imbalanced with the more Acute Lymphocytic Leukemia (ALL) class, the class weight method was used to balance the weight of two classes in pre-processing. Class weight function was followed because it is an optional dictionary mapping class, mostly used for loss function during training. It is one of the most efficient techniques to deal with an imbalanced dataset. From Sklearn, a compute_class_weight function with parameters of "balanced" was used.

$$Class\_weight = n\_samples/(n\_classes * np.bincount(y)) \qquad (1)$$

For the used dataset, the class weight of Normal Cell was 1.57288, and the class weight of Acute Lymphocytic Leukemia (ALL) cell was 0.73301.

{0: 1.5728828562997934, 1: 0.7330170517051705}

**Figure 2:** Class_weight

Then, different pre-trained models, namely InceptionV3, ResNet101V2, VGG19, InceptionResNetV2, were used to train the model. For further validation and explanation of the model, Local Interpretable Model-agnostic Explanation (LIME) algorithm was used, where the model indicates the focus point in a sample cell.

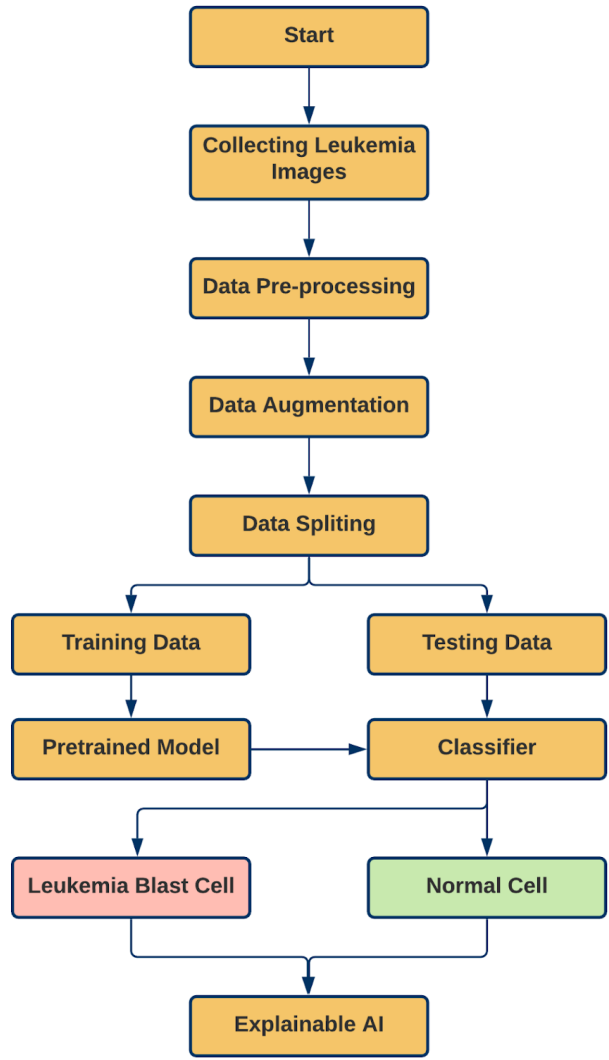

**Figure 3:** Block Diagram Of The Proposed Method

### *2.3 Data Pre-Processing*

All the images were reduced to 299px * 299px. Because the InceptionV3 input shape must be (299, 299, 3) and a standard batch size of 32 is used while training. For RGB images, the channel size should be 3. Up to this point, the images were not binary labeled, so Acute Lymphocytic Leukemia (ALL) cells and normal cells were labeled 1 and 0 correspondingly. By labeling the dataset, it is easy to understand, and all the data are simplified to a standard value.

*2.4 Data Splitting*

It can be difficult to assess a machine learning model. Typically, the data set is usually partitioned into testing and training sets. The model is trained using a training set, and model testing is performed using the testing set. Then, the correctness of the model is determined by evaluating its performance using an error metric. On the other hand, in a traditional method, the accuracy gained for one test set can be substantially different from another one. K-fold cross-validation offers a solution to this challenge. It separates the data into folds to ensure that each fold is used as a testing set.

*2.5 Data Augmentation*

The proposed method uses TensorFlow data augmentation functions such as random flip (up_down and also left_right). While training an image, it is important to use augmentation so that the model can identify a wide range of samples in real-life scenarios. The functions random_flip_up_down() and random_flip_left_right() randomly flip images vertically and horizontally, so that even if a leukemia-affected cell does not fit the training dataset, this model can still make a correct prediction in practice. A transpose is performed based on the spatial relationship of one image.tf.random.uniform() takes the datatype as float32, and if the spatial value is greater than 0.75, then it will perform a transpose. Also, depending on pixel size, saturation, contrast, brightness, and contrast are set. Some sample images may be brighter or more saturated than the training images in the test set. So it is highly beneficial to perform this augmentation of data. For the better focus of the model in the cell, some images underwent cropping functions for better effectiveness, but all images were later resized and reshaped to fit the model. As the images were in bitmap image file (bmp) format, they had to be decoded and converted to the tensor format. Tensor is a multidimensional array. Tensors can represent an image or video as an array. This helps the model to read the image easily. Finally, cancel the data duplication and sorting to ensure that the training phase is completely unbiased. If the model is trained using the same image multiple times, the model might be biased to predict test examples according to the repeated data image. Images in all the batches were first converted to a NumPy array along with their labels. The augmentation batch size is the same as the training batch size. Finally, the images were plotted using the matplotlib library to verify the process after augmentation. **Fig. 4** shows the output of dataset images after the augmentation process.

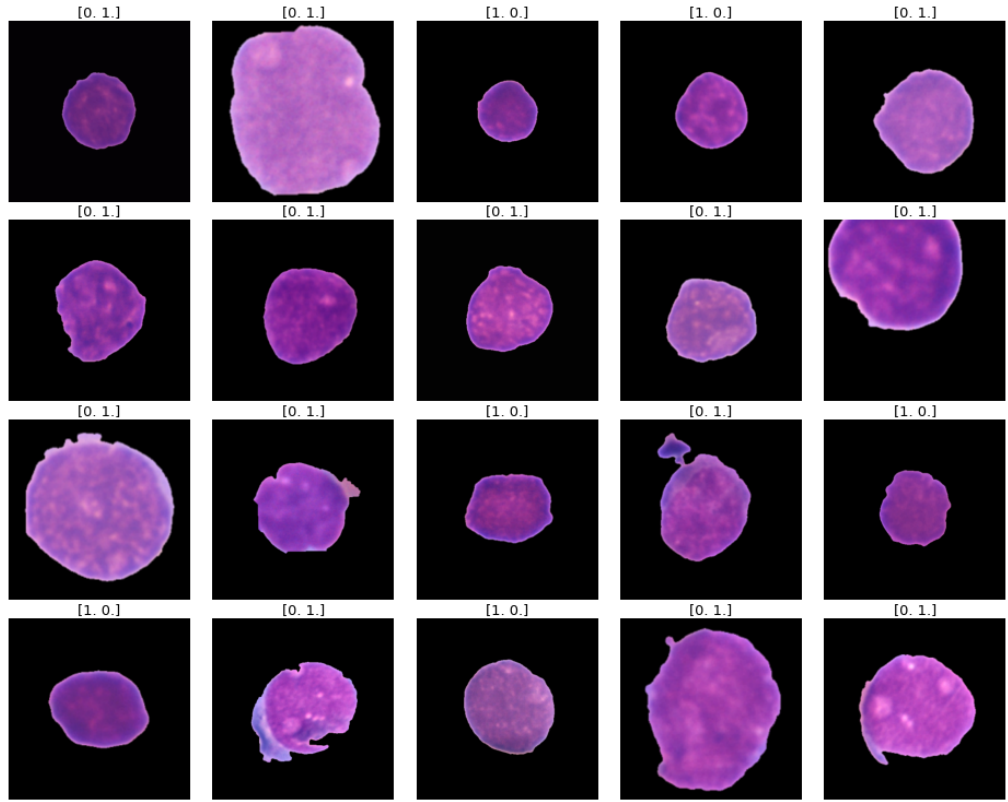

**Figure 4:** Augmented sample images

## *2.6 Convolutional Neural Network (CNN)*

A convolutional neural network (CNN) is a part of a deep learning method that takes an image sample as input and assigns priority to different neural in the sample image and distinguishes between distinct elements. The quantity of preprocessing required by CNN is much smaller than that required by other classification algorithms. While other basic techniques need hand-engineered filters, CNN can acquire these features appropriately. [17]

Using convolutional neural networks has been very successful in the case of image classification. A CNN's strength is its ability to automatically extract high-level information.First, the network architecture needs to be designed before training a CNN for image classification. This task entails determining the network's layer types, numbers, and order. The suggested network seeks to identify features to be utilized for differentiating classes with a set of 2D images and their accompanying class labels. A CNN learns using two repeated and alternated passes, called the "feedforward and backward pass" method. The feedforward pass accomplishes two significant tasks. The primary task is to extract features using many convolutional feature extraction (CFE) layers. For this reason, images are routed serially through many CFE layers. A CFE layer comprises three sublayers: a convolutional layer, a nonlinear transformation layer, and a pooling layer. Each CFE layer creates higher-level features by using features from the previous layer. Extracting advanced information from an image requires frequent repetition of this method. In the second task of the feedforward pass, the fully connected layers use these features to classify the sample image and obtain few errors. The feedforward pass propagates backward previous errors in a backward pass for altering the weights in the convolutional sublayers and allows extraction of more information concerning the classification problem. [18]

InceptionV3 is an extended version of the well-known GoogLeNet, which has shown high classification performance in various biomedical applications using transfer learning. Inception-v3 created an inception model that combines many convolutional filters of various sizes into a single new filter, similar to GoogLeNet. Due to this architecture, the number of training parameters is reduced. Hence, the computational complexity is reduced. The basic architecture of InceptionV3 is demonstrated in **Fig. 5**. [19]

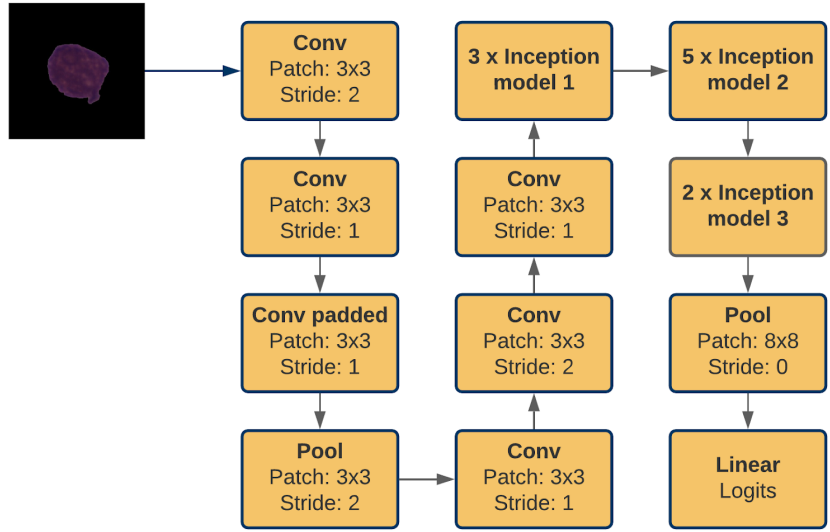

**Figure 5:** Architecture of InceptionV3

## *2.7 Transfer Learning*

A range of applications use deep convolutional neural networks because they can learn rich visual representations. However, for medical image-related problems, this needs a huge amount of data to complete the feature extraction. The dataset used for leukemia classification is insufficient to achieve good precision, resulting in overfitting the model. To overcome this problem, a transfer learning technique is used in this paper. Even if the data set is limited, transfer learning can improve model learning performance by solving the problem of insufficient samples. Transfer learning is a very effective but simple method to improve a network by transferring parameters from one domain (the target domain) to another domain. Transfer learning is a very effective but simple method to improve a network by transferring parameters from one domain (the target domain) to another domain. [20] **Fig. 6** shows the working process of conventional machine learning and transfer learning. In conventional machine learning, the model is trained from scratch, so the model requires more data to achieve a high score in performance matrices. But on the other hand, in the transfer learning technique, the model already has knowledge from the source task, so it requires very little data for the target task to get high scores in performance matrices.

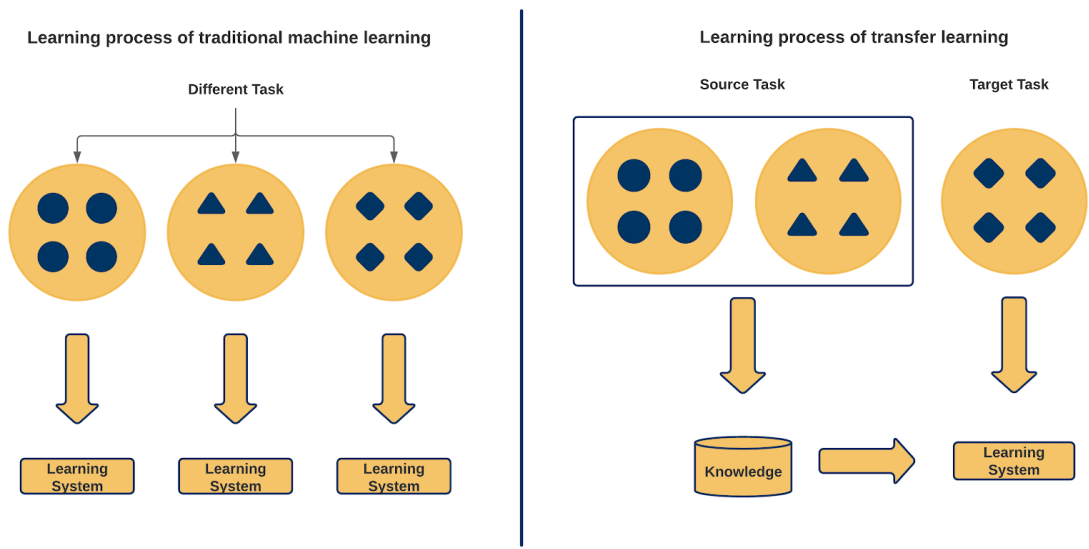

**Figure 6:** Difference between traditional machine learning and transfer learning

Adjusting the weight of data in the source domain is essential for usage in the target domain in a discriminatory manner. Transfer learning can outperform the scratch network since the pre-trained model already has a lot of basic information. The transfer learning method learns from the transferred domain about both low-end and mid-level properties. A modest amount of data from the new domain is required to achieve better outcomes. which is desirable for this work. [21]

The training process for the model was done in batches, with a set batch size of 32. Training in batches allows for computational speedup. Without splitting into batches, the deep learning algorithm has to store all the error values from 15,114 photos of the dataset in memory. Before training, all the batch images were converted into NumPy arrays. A transfer learning technique is applied in this research as it helps to get better accuracy using fewer data. In training, some well-known pre-trained models have been used. Machine learning models frequently fail to generalize properly when applied to data that has not been trained on. It occasionally fails miserably, and at other times it performs only marginally better than abysmal. We used a resampling technique called cross-validation to ensure that the model would perform well on unknown data. The entire data set is split up into k sets of about similar sizes in this resampling procedure. The model has been trained using the remaining k-1 sets, with the first set serving as the test set. The test error rate is computed after fitting the model to the test data. The second iteration uses the second set as a test set, whereas the remaining k-1 sets are utilized for training the data and determining the error. This method is repeated for all k sets. Stratified k-fold CV is used mostly in the event of classification difficulties, including class imbalance. In each training and validation fold, it ensures that the relative class proportion is nearly kept. In circumstances where there is a significant class divide, it becomes critical. Stratified K-fold cross-validation takes less time to compute and has a lower variance than traditional K-fold cross-validation. Furthermore, because more data points are used for invalidation, the MSE (Mean Square Error) will be less variable (variance).

*2.8 Pre-Trained Models*

One of the major problems in the field of medical research is the lack of data. But this problem can be overcome using transfer learning. The transfer learning technique minimizes the need for a large dataset by transferring the knowledge from a pre-trained model to a new model. The pre-train model consists of trainable and non-trainable layers. While training the pre-trained model using a new dataset, the trainable initial layers are replaced by the new layer.

It is necessary to verify the history of the training session in order to test the model. For this TensorFlow, the plot_metrics(history) function was used. The values of training loss, training accuracy, training f1 score, validation loss, validation accuracy, and validation f1 score are provided by the function history.keys().

The accuracy classifier is evaluated with an accuracy metric. The total number of data that provides accuracy is divided by the number of correctly classified data. Different values, such as True Positive (TP), False Positive (FP), True Negative (TN), and False Negative (FN) have been used to estimate the accuracy of this study.

$$Accuracy = \frac{TP+TN}{TP+TN+FP+FN} \qquad (2)$$

In a nutshell, validation and training accuracy indicate how well the model performed during the training and validation phases. The loss function depends on the formula that was followed to calculate the loss in the training or validation phase. The few common ways to calculate loss are binary cross-entropy, squared error loss, and absolute error loss. In the proposed method, the binary cross-entropy method was employed. The negative average of the log of the correct predicted probability is called binary cross-entropy.

$$logloss = -\frac{1}{N}\sum_i^N \sum_j^M y_{ij} \log(p_{ij}) \qquad (3)$$

Where, N is the number of rows

M is the number of classes

F1 score is used to measure the accuracy of a model on a dataset. It's determined by the calculated mean of recall and precision.

$$F1 - Score = 2 \times \frac{Recall \times Precision}{Recall + Precision} \qquad (4)$$

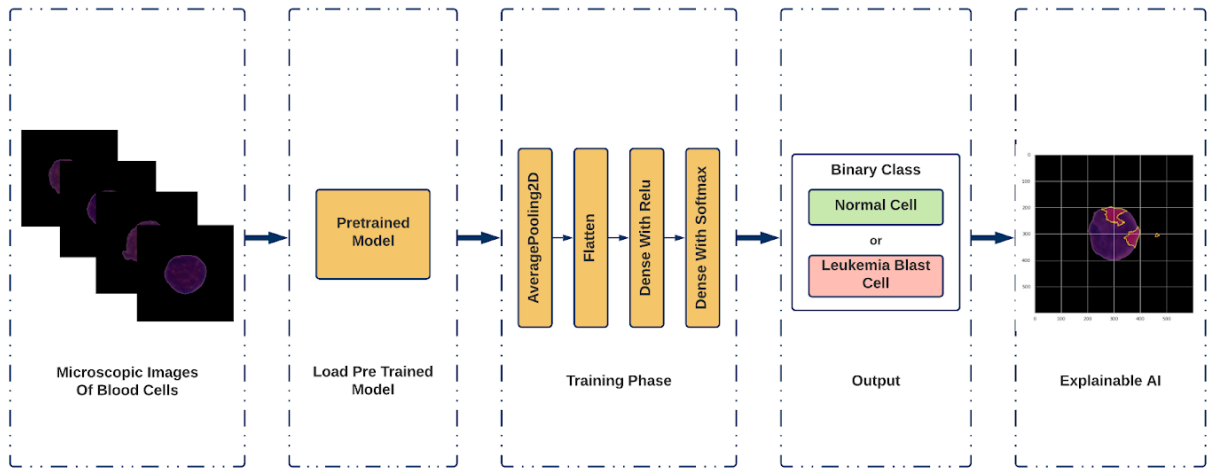

**Figure 7:** System architecture

## 2.9 Explainable AI

In the healthcare area, the ethical issue of AI transparency and a lack of trust in AI systems' black-box functioning demand the use of explainable AI models. Explainable AI (XAI) methods are AI methods that are used to explain AI models and their predictions. [22]

Explainable AI is one of the preeminent prerequisites for implementing responsible AI, a methodology for deploying AI approaches in real life while maintaining model explainability and responsibility. Ethical standards must be embedded in AI applications and processes to use AI responsibly, and AI systems must be built on trust and transparency.

Two research areas are particularly active in addressing this problem: the Explainable AI (XAI) field and the visual analytics community. Conversely, visual analytics solutions are designed to assist users in understanding and interacting with machine learning models by offering visualizations and tools that make exploring, analyzing, interacting with, and comprehending machine learning models easier. As a result, collaboration between the visual analytics and XAI communities is becoming increasingly vital.

**3 Results and Analysis**

After completing the data augmentation process, the models are trained using the Kaggle platform. The Kaggle kernel consists of an Nvidia P100 GPU, which has 16 gigabytes of GPU memory and 12 gigabytes of RAM. Each model has been trained with 35 epochs for each fold while training using the cross-validation technique. Adam is implemented as an optimizer, and Binary Cross Entropy is employed for model training as a loss function.

As the dataset has an imbalanced class distribution, every model is trained using stratified k-fold cross-validation. The whole dataset is split three-fold, and the 2nd fold performance matrix score is taken for each of the models. The score is given in **Tab. 1**

**Table 1:** Model accuracy, loss and F1-score

| Model | Accuracy | Loss | Validation Accuracy | Validation Loss | F1 Score | Validation F1 Score |
|---|---|---|---|---|---|---|
| ResNet101V2 | 0.9861 | 0.0333 | 0.9589 | 0.1559 | 0.9861 | 0.9588 |
| VGG19 | 0.9614 | 0.1060 | 0.9488 | 0.1425 | 0.9615 | 0.9487 |
| InceptionResNetV2 | 0.9914 | 0.0278 | 0.9564 | 0.1642 | 0.9914 | 0.9563 |
| InceptionV3 | 0.9838 | 0.0433 | 0.9665 | 0.1048 | 0.9839 | 0.9665 |

After observing the table, it can be stated that InceptionResNetV2 achieved train accuracy of 99.14%, which is the highest train accuracy compared to other trained models, and validation accuracy of 95.64%. ResNet101V2 and InceptionV3 also performed well. InceptionV3 achieved train accuracy of 98.38% and validation accuracy of 96.65%, the highest validation accuracy compared to other trained models. ResNet101V2 achieved train accuracy of 98.61% and validation accuracy of 95.89%. VGG19 achieved train accuracy of 96.14% and validation accuracy of 94.88%, the lowest among all trained models.

InceptionResNetV2 achieved a train loss of 2.78%, the lowest compared to other trained models. InceptionV3 achieved a train loss of 4.33%. However, InceptionV3 achieved a lower validation loss than InceptionResNetV2. InceptionV3 had a validation loss of 10.48%, while InceptionResNetV2 had a validation loss of 16.42%. ResNet101V2 achieved a train loss of 3.33% and validation loss of 15.59%. VGG19 achieved a train loss of 10.60%, the highest compared to other trained models, and a validation loss of 14.25%, lower than ResNet101V2.

In medical image classification, test set accuracy is important as test set accuracy is evaluated by the performance in the unknown dataset. According to **Tab. 2**, InceptionV3 achieved the highest test set accuracy. InceptionResNetV2 achieved a test set accuracy of 80.02%.

**Table 2:** Test set Accuracy and F1 Score

| Model | Test set accuracy | Test set F1 Score |
|---|---|---|
| ResNet101V2 | 0.7826 | 0.7770 |
| VGG19 | 0.7788 | 0.7695 |
| InceptionResNetV2 | 0.8002 | 0.7980 |
| InceptionV3 | 0.7981 | 0.7955 |

The entire dataset is divided into three folds since every model uses stratified k-fold cross validation. Hence, the model is trained for three iterations. The plot_metrics() function generates graphs per iteration of train accuracy, loss and f1 score for each of the models.

While training using ResNet101V2, the training accuracy has increased rapidly after each epoch. According to the accuracy and loss graph of ResNet101V2 shown in **Fig. 8**, the train accuracy was 50.00% in the first epoch, then increased with each epoch. After 20 epochs, the train accuracy is 95.52%. The model's validation accuracy was 43.56% in the first epoch, and it continued to increase until the last epoch when it achieved 95.89% in epoch 35. The model loss graph shows that both the training and validation loss line have gradually decreased. The training loss was 74.99% after the first epoch and 3.33% after 35 epochs.

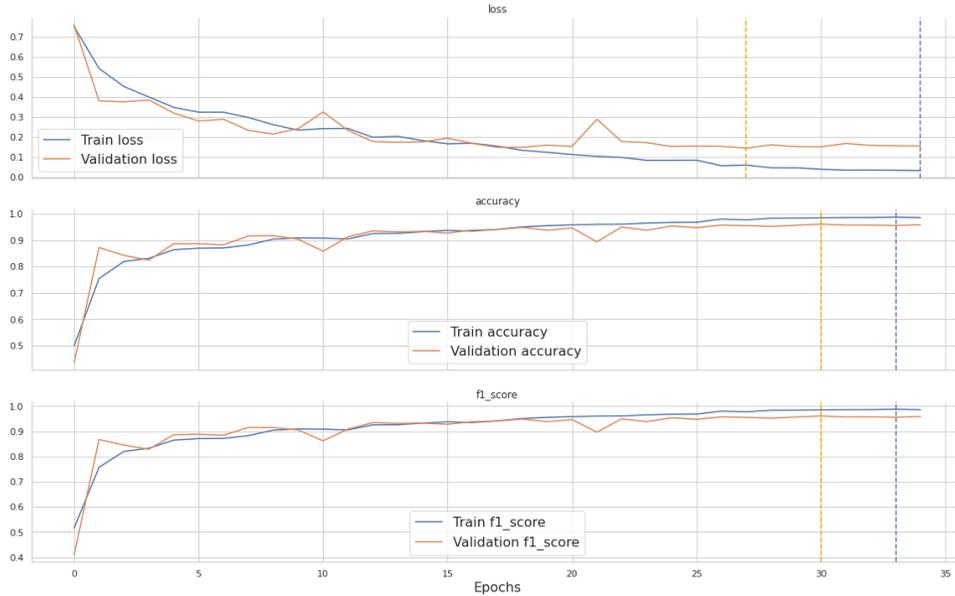

**Figure 8:** Accuracy, loss and F1-score of ResNet101V2

While training using VGG19, the training accuracy has increased rapidly after each epoch. According to the accuracy and loss graph of VGG19 shown in **Fig. 9**, the train accuracy was 64.63% in the first epoch, then increased with each epoch. After 20 epochs, train accuracy is 91.98%. The model's validation accuracy was 68.20% in the first epoch, and it continued to increase until the last epoch when it achieved 94.88% in epoch 35. The model loss graph shows that both the training and validation loss line have gradually decreased. The training loss was 76.66% after the first epoch and 10.60% after 35 epochs.

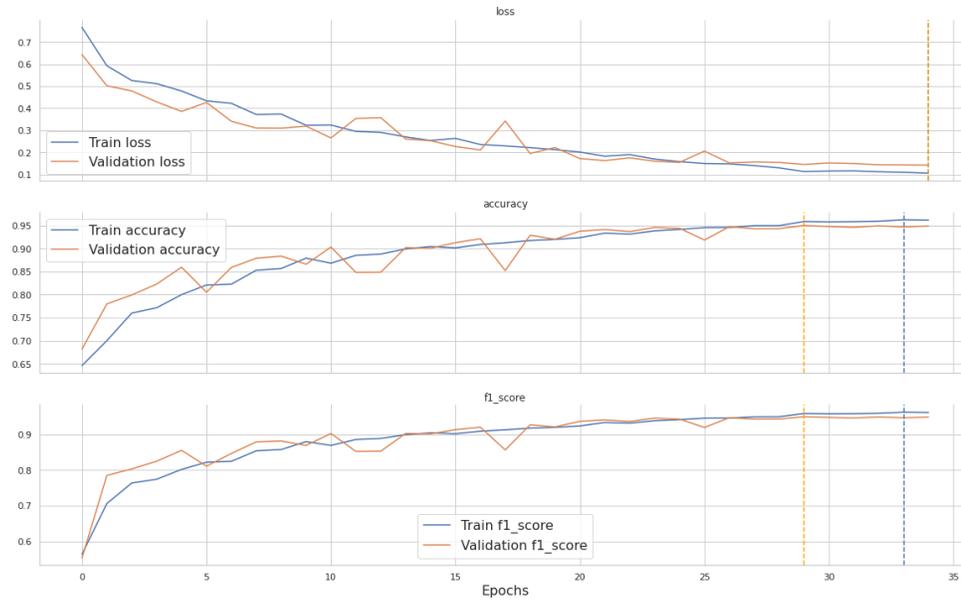

**Figure 9:** Accuracy, loss and F1-score of VGG19

While training using InceptionResNetV2, the training accuracy has increased rapidly after each epoch. According to the accuracy and loss graph of InceptionResNetV2 shown in **Fig. 10**, the train accuracy was 35.40% in the first epoch, then increased with each epoch. After 20 epochs, train accuracy is 97.47%. The model's validation accuracy was 35.42% in the first epoch, and it continued to increase until the last epoch when it achieved 95.64% in epoch 35. The model loss graph shows that both the training and validation loss line have gradually decreased. The training loss was 74.44% after the first epoch and 2.78% after 35 epochs.

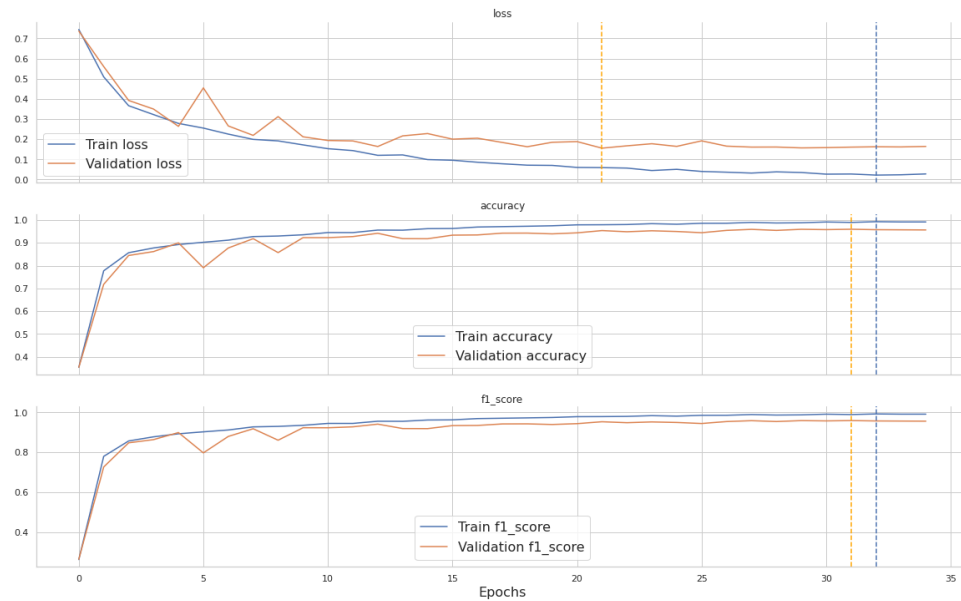

**Figure 10:** Accuracy, loss and F1-score of InceptionResnetV2

While training using InceptionV3, the training accuracy has increased rapidly after each epoch. According to the accuracy and loss graph of InceptionV3 shown in **Fig. 11**, the train accuracy was

39.06% in the first epoch, then increased with each epoch. After 20 epochs, train accuracy is 98.82%. The model's validation accuracy was 35.23% in the first epoch, and it continued to increase until the last epoch when it achieved 96.65% in epoch 35. The model loss graph shows that both the training and validation loss line have gradually decreased. The training loss was 72.79% after the first epoch and 4.33% after 35 epochs.

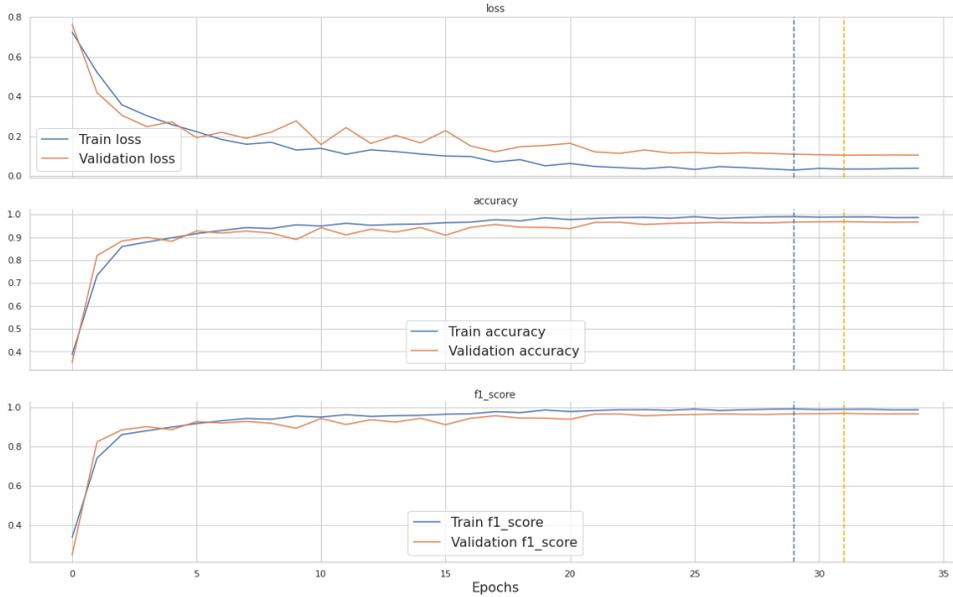

**Figure 11:** Accuracy, loss and F1-score of InceptionV3

Among all the models, InceptionV3 gave the clearest BlackBox explanation. That is why the InceptionV3 model was picked to identify leukemia cells. 'ImageNet' weight was preferred for pre-trained model weight, as it has the most robust training. InceptionV3 takes Average pooling then flattens the array. Finally, using softmax from the dense layer sample images was predicted as a normal cell or leukemia blast cell.

In the proposed method, LIME (Local Interpretable Model-agnostic Explanation) is applied and visual interpretation is used for describing the model. LIME is a model-agnostic algorithm that approximates the local linear behavior of the model, which means it can explain any model of CNN and NLP. [23] An explanation has to be presented so that it is easily understandable by a human being.

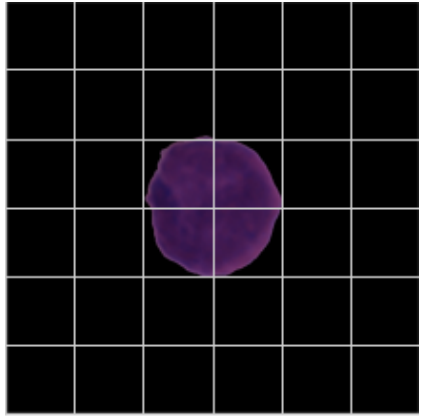

**Figure 12:** Sample Cell

Fig. 12 is a sample taken to predict and later explain with the LIME algorithm. The model predicted that this sample image would be ALL. For prediction, the proposed model takes the highest value as a classification result. In the case of this sample, the maximum value was 0.99 towards ALL. So, the prediction was an ALL image.

```
The Prediction of the sample is: It Is ALL
Prediction Confidence Percentage is:  99.99970197677612
```

**Figure 13:** Prediction result of a given sample

A segmentation method was employed to divide the example image into separate sections to see if the model could accurately read it. It also tells us what the sample image represents to the model. By observing the segmentation result, one can easily tell the separation between the foreground and background of the sample image. **Fig. 14** shows the sample cell is in the middle with a circular shape highlighted in red.

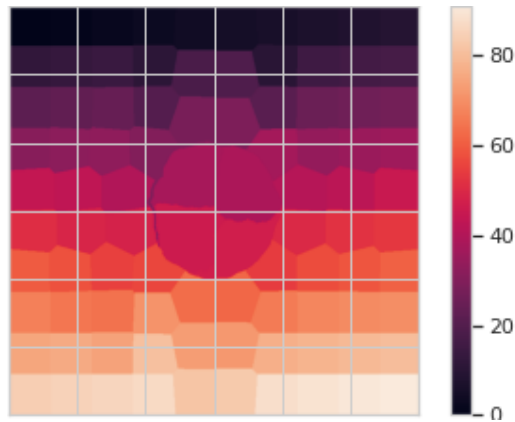

**Figure 14:** Model Segmentation

After that, the 3D image boundary needed to be verified to understand the model's reliability. A scikit-learn image segmentation function was carried out for the 3D boundary. This returns the sample image with its boundaries highlighted between labeled regions. In **Fig. 15**, the boundary is clearly labeled around the sample image to create a 3D depth around the cell. This represents the model's awareness that the intended target is correct.

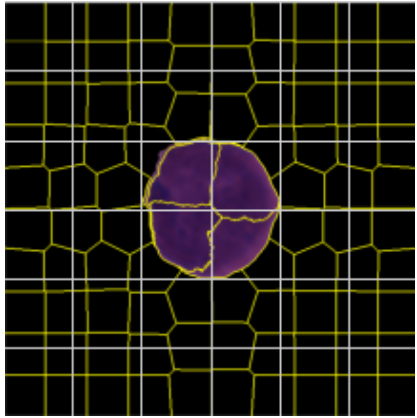

**Figure 15:** Model Boundaries

As the LIME algorithm is model-agnostic and can explain both classification and regression models, the proposed model uses the LIME image explainer. For this package to work, the image has to be a 3D NumPy array. It explains image prediction by sampling from 0 and then inverting the mean-centering and scaling operations. As our model solves a classification problem, it samples the training distribution, and when the value is the same, it makes a binary feature that is 1. After the explainer is set, the instance from lime_image generates neighborhood data. After learning the weight of the linear models locally, explanations can be extracted from the model. Top_labels shows the highest weight of the prediction probability considered for that particular sample image. Finally, the model can explain the major weight behind any prediction. For the above sample image, the model puts more weight on that which is highlighted in red, as shown in **Fig. 16**.

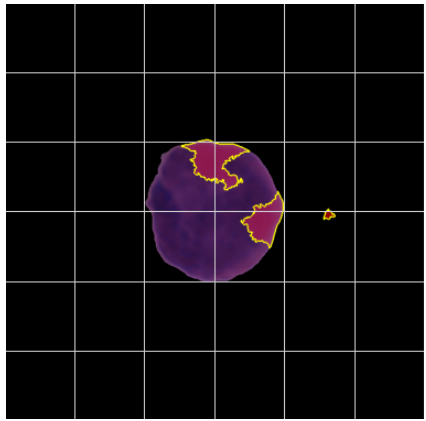

**Figure 16:** Image Temperature

By highlighting this, any doctor can verify if the model was right in predicting all of this sample image. Also, in **Fig. 17**, only the weighted part is made more prominent by isolation. This makes the image's explanation more understandable.

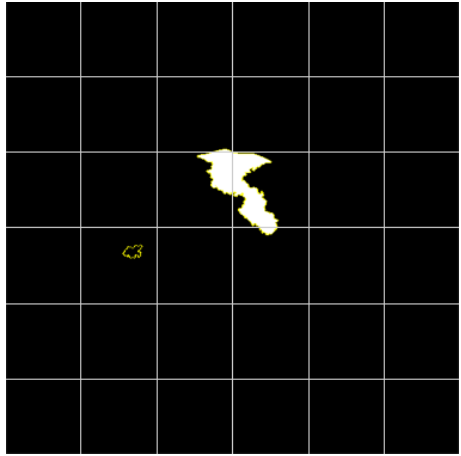

**Figure 17:** Image Temperature (positive only)

The accuracy of the proposed model (InceptionV3) was 98.38%, and the F1 score was 98.39%. The training loss in the final epoch for the model was 4.33%, and the validation loss was 10.48% with the most explicit BlackBox explanation. Moreover, InceptionV3 achieved the highest validation accuracy and f1-score with lowest validation loss compared to all other trained models.

The best trained models used in this paper were compared to the papers mentioned above. The accuracy is given in **Tab. 3** With the help of Explainable AI, the accuracy of all our trained models is sufficient to diagnose leukemia with ease.

**Table 3:** Result Comparison With Previous Work

| Paper | Accuracy | This Paper Accuracy |
| --- | --- | --- |
| In paper [5] | 80.40% | |
| In paper [6] | 96.58% | |
| In paper [8] | 88.69% | 98.38% |
| In paper [9] | 96.15% | |
| In paper [10] | 99.03% | |
| In paper [11] | 96.42% | |

On account of train accuracy, InceptionV3 has the second-highest score but has the highest validation accuracy and validation f1-score. Furthermore, the InceptionV3 model provided the best fit in terms of explainable AI, which is more important in medical fields. That is why, for the described method, InceptionV3 was preferred above the rest of the compared models.

## 4 Conclusion

This paper proposed a novel and efficient system, a diagnostic approach for Acute lymphocytic Leukemia (ALL), that compares different transfer learning models to identify malignant cells and normal cells to assist doctors in acute lymphocytic leukemia diagnosis. The proposed system provides 98.38% accuracy in diagnosing acute lymphoblastic leukemia in patients. The result indicates that the model provided more accurate results. By comparing different transfer learning models, they had a more balanced classification capacity that could give a kickstart for using different transfer learning models to diagnose acute lymphoblastic leukemia. This method also uses Local Interpretable Model-Agnostic

Explanations (LIME) to describe which component of the image from the dataset caused the model to produce specific classifications, ensuring the model's validity and reliability. Therefore, the proposed approach gives clinicians a reliable way to diagnose whether or not a patient has leukemia. This system can be used to make an initial ALL diagnosis, after which further testing can be done. The method proposed in this study is a highly promising methodology to identify ALL. In the future, advanced work will dramatically improve the overall state of this system and its explanation. It will provide behavioral inferences and useful insights into deep network operations. This will build trust in deep learning systems as well as allow for system behavior understanding and improvement.

**Data Availability:** The data used to assist the finding of this study are openly accessible at:
Link: https://www.kaggle.com/andrewmvd/leukemia-classification

**Conflicts of Interest:** The authors declare no conflicting interests to disclose regarding this work.

**Acknowledgement:** The authors are grateful to Dr. Mohammad Monirujjaman Khan, Department of Electrical and Computer Engineering, North South University, Dhaka, Bangladesh, for his generous support in conducting this research. The authors also convey their appreciation to the reviewers in advance for their thoughts and recommendations.